%% file: main.tex
\def\year{2022}\relax
\documentclass[letterpaper]{article} 
\usepackage{aaai22}  
\usepackage{times}  
\usepackage{helvet}  
\usepackage{courier}  
\usepackage[hyphens]{url}  
\usepackage{graphicx} 
\usepackage{natbib}  
\usepackage{caption} 
\usepackage[utf8]{inputenc}
\usepackage[flushleft]{threeparttable}
\usepackage{multirow}
\usepackage{booktabs}
\usepackage{amsmath,amssymb}
\usepackage{xspace}
\usepackage[english]{babel}
\usepackage{graphicx}
\usepackage{floatrow}
\usepackage{xcolor}
\usepackage{comment}
\usepackage{tikz}
\usepackage{graphicx} 
\usepackage{pgfplots}
\usepackage{subcaption}
\usepackage{ntheorem}
\graphicspath{ {./figures/} }

\DeclareCaptionStyle{ruled}{labelfont=normalfont,labelsep=colon,strut=off} 
\usepackage{algorithm}
\usepackage{algorithmic}

\input{define}
\title{\method: Temporal Question Reasoning over Knowledge Graphs}
\author{Costas Mavromatis\textsuperscript{\rm 1}\thanks{Corresponding Author: \{mavro016@umn.edu\}. Work done while interning at Amazon Web Services.}\equalcontrib, Prasanna Lakkur Subramanyam\textsuperscript{\rm 2}\thanks{Work done while interning at Intel.}\equalcontrib, Vassilis N. Ioannidis\textsuperscript{\rm 3}, Soji Adeshina\textsuperscript{\rm 3}, Phillip R. Howard\textsuperscript{\rm 4}, Tetiana Grinberg\textsuperscript{\rm 4}, Nagib Hakim\textsuperscript{\rm 4}, George Karypis\textsuperscript{\rm 3}}
\affiliations{

    \textsuperscript{\rm 1}University of Minnesota,
    \textsuperscript{\rm 2}University of Massachusetts Amherst,
    \textsuperscript{\rm 3}Amazon Web Services,
    \textsuperscript{\rm 4}Intel Labs\\
}

\begin{document}

\maketitle

\begin{abstract}
Knowledge Graph Question Answering (KGQA) involves retrieving facts from a Knowledge Graph (KG) using natural language queries. A KG is a curated set of facts consisting of entities linked by relations. Certain facts include also temporal information forming a Temporal KG (TKG). Although many natural questions involve explicit or implicit time constraints, question answering (QA) over TKGs has been a relatively unexplored area. 
Existing solutions are mainly designed for simple temporal questions that can be answered directly by a single TKG fact.
This paper puts forth a comprehensive embedding-based framework for answering complex questions over TKGs. Our method, called temporal question reasoning~(\method), exploits TKG embeddings to ground the question to the specific entities and time scope it refers to. It does so by augmenting the question embeddings with context, entity and time-aware information via three specialized modules. The first computes a textual representation of a given question, the second combines it with the entity embeddings for entities involved in the question, and the third generates question-specific time embeddings. Finally, a transformer-based encoder learns to fuse the generated temporal information with the question representation, which is used for answer predictions.
Extensive experiments show that \method improves accuracy by 25--45 percentage points on complex temporal questions over state-of-the-art approaches and it generalizes better to unseen question types. 
\end{abstract}
\section{Introduction}

A knowledge graph (KG) is a set of facts that are known to be true in the world or in a specific domain. The facts are usually represented as tuples (subject, relation, object), where the subject and object correspond to KG entities. Certain KGs include additional attributes such as temporal information forming a temporal knowledge graph (TKG). In TKGs, each fact is associated with a timestamp or time interval, and is represented as (subject, relation, object, timestamp) or  (subject, relation, object, [start\_time, end\_time]), respectively. 



Knowledge Graph Question Answering (KGQA) attempts to answer a natural question using the KG as a knowledge base~\cite{lan2021qasurvey}. Natural questions often include temporal constraints, e.g., \textit{``Which movie won the Best Picture in 1973?"} and to aid temporal question answering, TKGs are utilized. The first step is to identify and link the entities,  relations and timestamps of the questions to the corresponding ones in the TKG, e.g., ``\textit{Which movie won the Best Picture in 1973?}" to (Best Picture, WonBy, ?, 1973). This problem is known as entity linking~\cite{kolitsas2018endtoend}. 

Recently,~\cite{saxena2021cronkgqa} proposed CronKGQA that solves QA over TKGs by leveraging TKG embedding methods, e.g., TComplEx~\cite{Lacroix2020Tcomplex}.  TKG embedding methods learn low-dimensional embeddings for the entities, relations and timestamps by minimizing a link prediction objective attuned at completing facts of the form (subject, relation, ?, timestamps) and (subject, relation, object, ?). CronKGQA answers the mapped question (Best Picture, WonBy, ?, 1973) as a link prediction task over the TKG. 
CronKGQA performs very well on simple questions that are answerable by a single TKG fact (Hits@1 of 0.988). However, on more complex questions, that involve additional temporal constraints and  require information from multiple TKG facts (e.g., ``\textit{Which movie won the Best Picture after The Godfather?}"), CronKGQA performs poorly (Hits@1 of 0.392).

\begin{figure*}
  \centering
  \begin{subfigure}[t]{0.45\columnwidth}
    \resizebox{.95\linewidth}{!}{\input{figures/framework/phase1}}
    \caption{The underlying TKG.}
    \end{subfigure}
    \hfill
    \begin{subfigure}[t]{0.45\columnwidth}
        \resizebox{.95\linewidth}{!}{\input{figures/framework/phase2}}
        \caption{Context-aware question representation.}
    \end{subfigure}
    \hfill
    \begin{subfigure}[t]{0.45\columnwidth}
        \resizebox{.95\linewidth}{!}{\input{figures/framework/phase3}}
        \caption{Entity-aware question representation.}
    \end{subfigure}
    \hfill
    \begin{subfigure}[t]{0.45\columnwidth}
        \resizebox{.95\linewidth}{!}{\input{figures/framework/phase4}}
        \caption{Time-aware question representation.}
    \end{subfigure}
    \caption{(a) The underlying TKG for the question ``\textit{Which movie won the Best Picture after The Godfather?}"; Answer: `The Sting'.  (b) The context-aware question representation (dashed arrows) scores higher (numbers in parentheses) movie entities. (c) The representation is grounded to the entity `Best Picture' and gives higher scores to movies that won the Best Picture. (d) The representation is grounded to the correct time scope (after year 1972) and gives higher scores to the movies that won the Best Picture after 1972, i.e., `The Sting'.}
    \label{fig:framework}
\end{figure*}
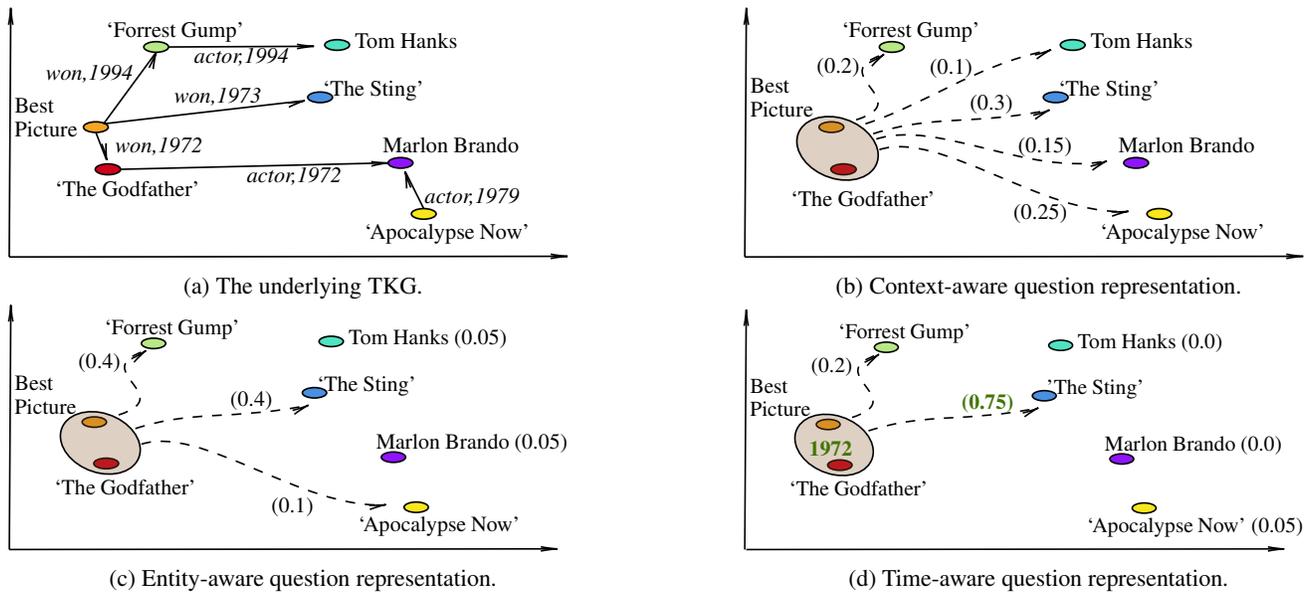 

To effectively handle both simple and complex temporal questions, we design a new method called temporal question reasoning (\method). \method exploits TKG embeddings to ground the question to the specific entities and time scope the question refers to. 
To illustrate the key idea of our approach, consider the question ``\textit{Which movie won the Best Picture after The Godfather?}" which involves the TKG entities `Best Picture' and `The Godfather'.
The high-level approach of \method is shown in Figure~\ref{fig:framework}.
The first reasoning step is to understand the context of the question (context-aware step). The context of the question here involves a movie (\textit{``Which movie..."}). The next step is to ground the question to the entities it refers to (entity-aware step). The question refers to a movie that has won the Best Picture. Finally, the question needs to be grounded with certain temporal constraints (time-aware step). The movie won the Best Picture after The Godfather, i.e., after 1972.

\method performs the above reasoning steps using three specialized modules. First, it 
uses the question's text to generate a representation of the question by employing a language model (LM). Next, it fuses the text-derived representations with KG entity representations to ground the question to the entities it refers to. Third, it extracts from the TKG question-specific temporal information to enhance the question's representation with two complementary approaches. The first \emph{retrieves} the relevant information from the underlying TKG  based on the annotated entities of the question. The second \emph{infers} temporal information by solving a link prediction problem obviating the need to access the TKG. A dedicated information fusion layer combines the context, entity and time-aware  information together to a final question representation. We empirically show that \method{} is able to model temporal constraints of the question and outperforms other time-unaware methods for complex questions (25--45 percentage points improvement at Hits@1). Our contributions are summarized below:
\begin{itemize}
    \item We solve complex temporal question answering by learning context, entity and time-aware question representations.
    \item We develop two different approaches to recover question-specific temporal information.
    \item We achieve state-of-the-art performance on temporal complex questions and provide additional strong baselines.
\end{itemize}



\section{Related Work} \label{sec:rel}

KGQA approaches typically leverage KG pre-trained embeddings~\cite{bordes2013translating,yang2014distmult,Trouillon2017Complex} to answer the questions~\cite{saxena2020embedkgqa}. Such approaches perform well for simple questions that can be easily mapped to incomplete facts in the KG but are challenged by complex questions. 


Addressing the limitations of the aforementioned approaches,~\cite{miller2016kvm,xu2019enhancing-kvm,Zhou2018multi-rel,Qiu2020stepwise,He2021intermediate} enhance the question representation to address complex questions. 
Such methods employ logical reasoning~\cite{miller2016kvm,xu2019enhancing-kvm,Zhou2018multi-rel,Qiu2020stepwise,He2021intermediate} or leverage available side information in the form of text documents~\cite{sun2018graftnet,Sun2019PullNet,xiong2019sgreader,Han2020HyperGraph}. 
Nevertheless, these approaches are not suited for handling temporal constraints.

TempQuestions~\cite{jia2018tempquestions} was introduced to benchmark the reasoning capabilities of existing with temporal constraints. Recently, additional benchmarks have been developed~\cite{jin2021forecastqa,souza2020event,chen2021dataset,neelam2021sygma} that model temporal information in various domains (including both KG and text data).
~\cite{jia2018tequila} and ~\cite{jia2021complex} are methods that tackle the temporal QA problem over KGs. However, they mostly employ hand crafted rules to handle temporal information which is not flexible for incomplete KGs. By leveraging TKG embeddings, CronKGQA~\cite{saxena2021cronkgqa} provides a learnable reasoning process for temporal KGQA, which does not rely on hand-crafted rules. Although CronKGQA performs extremely well for answering simple questions, it is challenged by complex questions that require inference of certain temporal constraints. Our work is motivated by this limitation.

\section{Background}

A TKG $\mathcal{K} := (\mathcal{E}, \mathcal{R}, \mathcal{T}, \mathcal{F})$ 
contains a set of entities $\mathcal{E}$, a set of relations $\mathcal{R}$, a set of timestamps $\mathcal{T}$, and a set of facts $\mathcal{F}$. Each fact $(s, r, o, \tau) \in \mathcal{F}$ is a tuple where  $s,o \in \mathcal{E}$ denote the subject and object entities, respectively, $r \in \mathcal{R}$ denotes the relation between them, and  $\tau \in \mathcal{T}$ is the timestamp associated with that relation. 


\subsection{TKG embeddings}



Given a TKG $\mathcal{K}= (\mathcal{E}, \mathcal{R}, \mathcal{T}, \mathcal{F})$, TKG embedding methods typically learn $\ndims$-dimensional vectors $\mathbf{e}_{\epsilon}, \mathbf{v}_r, \mathbf{t}_{\tau} \in \mathbb{R}^\ndims$ for each $\epsilon \in \mathcal{E}$, $r \in \mathcal{R}$ and $\tau \in \mathcal{T}$. 
These embedding vectors are learned such that each valid fact $(s,r,o,\tau) \in \mathcal{F}$ is scored higher than an invalid fact  $(s',r',o',\tau') \notin \mathcal{F}$ through a scoring function $\phi(\cdot)$, i.e., $\phi(\mathbf{e}_s,\mathbf{v}_r,\mathbf{e}_o,\mathbf{t}_\tau) > \phi(\mathbf{e}_{s'},\mathbf{v}_{r'},\mathbf{e}_{o'},\mathbf{t}_{\tau'})$. Please see \cite{kazemi2020tkge} for notable TKG embedding methods.






\noindent
\textbf{TComplEx}. TComplEx~\cite{Lacroix2020Tcomplex} is an extension of the ComplEx~\cite{Trouillon2017Complex} KG embedding method designed for TKGs. TComplEx represents the embeddings as complex vectors in $\mathbb{C}^{\ndims/2}$ and the scoring function $\phi(\cdot)$ is given by
\begin{align}
\begin{split}
    \phi(\mathbf{e}_s,\mathbf{v}_r,\bar{\mathbf{e}}_o,\mathbf{t}_\tau) = \text{Re}(\langle \mathbf{e}_s,\mathbf{v}_r \odot  \mathbf{t}_\tau ,\bar{\mathbf{e}}_o\rangle)
\end{split}
\label{eq:tcomplex}
\end{align}
where $\text{Re}(\cdot)$ denotes the real part, $\bar{(\cdot)}$ is the complex conjugate of the embedding vector and $\odot$ is the element-wise product.

TComplEx employs additional regularizers to improve the quality of the learned embeddings such as enforcing close timestamps to have similar embeddings (closeness of time). The embedding learning procedure makes TComplEx a suitable method for inferring missing facts such as $(s,r,?,t)$ and $(s,r,o,?)$ over an incomplete TKG. Throughout the manuscript, we generate TKG embeddings via TComplEx due to its aforementioned benefits.


\input{tables/cronqs}

\subsection{QA over TKGs} \label{sec:tkgqa}
Given a TKG $\mathcal{K} = (\mathcal{E}, \mathcal{R}, \mathcal{T}, \mathcal{F})$ and a natural language question $q$, the task of QA over a TKG (TKGQA)~\cite{saxena2021cronkgqa} is to extract an entity $ \epsilon' \in \mathcal{E}$ or  a timestamp $ \tau' \in \mathcal{T}$ that correctly answers the question $q$. The entities $ \epsilon \in \mathcal{E}$ and timestamps $\tau \in \mathcal{T}$ of the question are  annotated, i.e., linked to the TKG. Please refer to Table~\ref{tab:cronqs} for examples of such questions. 

Note that a question, e.g.,  \textit{``Which movie won the Best Picture in 1973"}, could be answerable by a single TKG fact, i.e., (Best Picture, WonBy, The Sting, 1973).  Thus, a common solution for TKGQA is to infer the relation `WonBy' by the question's context and solve the problem as link prediction, i.e., (Best Picture, $q$, ?, 1973). 

\noindent
\textbf{CronKGQA}. CronKGQA~\cite{saxena2021cronkgqa} is a typical method that solves TKGQA as a link prediction task. The idea is to use the question as a `virtual relation' in the scoring function $\phi(\cdot)$. 

Suppose $s$ and $\tau$ are respectively the annotated subject and timestamp in a given question (e.g., `Best Picture' and `1973') and $o^\ast$ is the correct answer (e.g., `The Sting'). 
CronKGQA learns a question representation $\mathbf{q}$ such that $\phi(\mathbf{e}_s, \mathbf{q}, \mathbf{e}_{o^\ast}, \mathbf{t}_\tau) > \phi(\mathbf{e}_s, \mathbf{q}, \mathbf{e}_{o'}, \mathbf{t}_\tau)$ for all incorrect entities $o'\ne o^\ast$, where $\mathbf{e}_s, \mathbf{e}_{o^\ast}, \mathbf{e}_{o'},$ and $\mathbf{t}_\tau$ are pre-trained TKG embeddings, e.g., with TComplEx.  Note that if either $s$ or $\tau$ is not present in the question, a random one from the TKG (dummy) is used. The methodology is modified accordingly when the answer is a timestamp $\tau^\ast$ by giving the maximum score to $\phi(\mathbf{e}_s, \mathbf{q}, \mathbf{e}_{o}, \mathbf{t}_{\tau^\ast})$, where $\tau^\ast$ is the correct timestamp and $s, o$ are the annotated subject and object.

CronKGQA is designed for simple temporal questions that can  be transformed to link predictions over TKGs. This limits the applicability of CronKGQA to more complex questions that involve additional temporal constraints, e.g., `Before/After', `First/Last' and `Time Join' questions of Table~\ref{tab:cronqs}. This is confirmed by the experiments presented in~\cite{saxena2021cronkgqa}, where CronKGQA achieves a  65\% performance improvement over non-TKG embedding methods for simple questions, but only a 15\% improvement for complex questions as the ones here.

\section{Method:~\method}
\method{} leverages pre-trained TKG embeddings of entities and timestamps that encode their temporal properties, i.e., TComplEx. Although TKG embeddings are designed for simple questions, our method overcomes this shortcoming by incorporating additional temporal information in the question representation  $\mathbf{q}$ to better handle constraints. We design \method{} by following the human reasoning steps to answer temporal questions; see also Figure~\ref{fig:framework}. In the following subsections, we describe in detail the architecture of \method.

\subsection{Context-aware question representation.}

Given a question's text, we use pre-trained LMs, e.g., \textsc{Bert}~\cite{devlin2019bert}, to encode the question's context into an embedding vector.  The [CLS] token is inserted into the question, e.g., ``[CLS] \textit{Which movie won the Best Picture after The Godfather?}", which is transformed to a tokenized vector $\mathbf{q}_0$. Then, we compute a representation for each token as
\begin{equation}
    \mathbf{Q}_B =  \mathbf{W}_B \textsc{BERT}(\mathbf{q}_0)  \;,
\label{eq:Q1}
\end{equation}
where $\mathbf{Q}_B:=[\mathbf{q}_{B_{\text{CLS}}}, \mathbf{q}_{B_1}, \ldots, \mathbf{q}_{B_N}]$ is a $\ndims\times\ntokens$  embedding matrix where $\ntokens$ is the number of tokens and $\ndims$ are the dimensions of the TKG embeddings.  $\mathbf{W}_B$ is a $\ndims\times\bndims$ learnable projection matrix where $\bndims$ is the output the dimension of the LM ($\bndims=768$ for  \textsc{BERT}($\cdot$)).


\subsection{Entity-aware question representation.}
We utilize the TKG entity embeddings to ground the question to the specific entities it involves. Inspired by other approaches~\cite{zhang2019ernie,Fvry2020EntitiesAE} that compute entity-aware text representations, we replace the token embeddings of the entities and timestamps of $\mathbf{Q}_B$ with their pre-trained TKG embeddings. Specifically, the $i$th column of the  entity-aware token embedding matrix $\mathbf{Q}_E$ is computed as
\begin{equation}
    \mathbf{q}_{E_i} = 
    \begin{cases}
    
     \mathbf{W}_E \mathbf{e}_{\epsilon} , & \text{if token $i$ is linked to an entity $\epsilon$,} \\
    \mathbf{W}_E  \mathbf{t}_{\tau}, & \text{if token $i$ is linked to a timestamp $\tau$.}\\
    \mathbf{q}_{B_i}, & \text{otherwise,}\\
\end{cases}
\end{equation}
where $\mathbf{W}_E$ is a $\ndims\times\ndims$ learnable projection.
As a result, the token embedding matrix $\mathbf{Q}_E:=[\mathbf{q}_{E_{\text{CLS}}}, \mathbf{q}_{E_1}, \ldots, \mathbf{q}_{E_N}]$ incorporates additional entity information from the TKG.  This enriches the question with entity information from the TKG. 

\subsection{Time-aware question representation.} \label{sec:TAQE}
A question may refer to a specific time scope which the answer needs to be associated, e.g., ``\textit{after The Godfather}" refers to ``\textit{after 1972}".
We develop two alternatives that recover such temporal information. The first approach \emph{retrieves} the question-specific time scope from the TKG based on the annotated entities. We call this approach \emph{hard-supervised}, since it accesses available facts in the TKG. The second approach \emph{infers} question-specific temporal information based on the question's representation. We term this approach \emph{soft-supervised}, since it may recover missing temporal facts by operating in the embedding space.


\noindent
\textbf{Hard Supervision: Retrieval from the TKG facts.}
We utilize the annotated entities of the question to retrieve the relative time scope from the underlying TKG. For the example question ``\textit{Which movie won the Best Picture after The Godfather?}", the  entities `Best Picture' and `The Godfather' appear together in a TKG fact with timestamp 1972. Hence, the time embedding of 1972 can be utilized to further enhance the question representation.

First, we identify all facts that involve the annotated entities of the question. These facts involve specific timestamps which we collect together (retrieved timestamps). 
In some cases, we may retrieve multiple timestamps, but we only keep the \emph{start} and \emph{end} timestamps after we sort them (since we aim at recovering a question-specific time scope). 
We recover two temporal embeddings $\mathbf{t}_1$ and $\mathbf{t}_2$ that correspond to the TKG embedding for start and end timestamps, respectively. We term this method  \methodh.

\noindent
\textbf{Soft Supervision: Inference in the TKG embedding space.}
Instead of retrieving timestamps from the TKG, we may directly obtain time embeddings by utilizing $\phi$ to infer missing temporal information. We generate a time-aware question embedding $\mathbf{q}_{\text{time}}$
as
\begin{equation}
    \mathbf{q}_{\text{time}} =  \mathbf{W}_T \mathbf{q}_{B_{\text{CLS}}} ,
\end{equation}
where $\mathbf{q}_{B_{\text{CLS}}}$ corresponds to the [CLS] token embedding of $\mathbf{Q}_B$ and $\mathbf{W}_T$ is a $\ndims\times\bndims$ learnable projection matrix. The time-aware $\mathbf{q}_{\text{time}}$ is used as a `virtual relation' in the scoring function $\phi$.
TComplEx assigns a score to a timestamp $\tau$ that potentially completes a fact $(s,r,o,?)$ as 
\begin{equation}
\begin{split}
    \big\langle \text{Re}(\mathbf{e}_s) \odot \text{Re}(\mathbf{u}_{ro}) - \text{Im}(\mathbf{e}_s) \odot \text{Im}(\mathbf{u}_{ro}),  \text{Re}(\mathbf{t}_\tau) \big\rangle & +\\
    \big\langle  \text{Re}(\mathbf{e}_s) \odot \text{Im}(\mathbf{u}_{ro}) + \text{Im}(\mathbf{e}_s) \odot \text{Re}(\mathbf{u}_{ro}),  \text{Im}(\mathbf{t}_\tau) \big\rangle &,
\end{split}
\end{equation}
where $\mathbf{u}_{ro} = \mathbf{v}_r \odot \bar{\mathbf{e}}_o$. Thus, the real \text{Re($\cdot$)} and imaginary \text{Im($\cdot$)} part of the time embedding $\mathbf{t}_\tau$ can be approximated by
\begin{equation}
\begin{split}
     \text{Re}(\mathbf{t}_\tau) \approx \text{Re}(\mathbf{e}_s) \odot \text{Re}(\mathbf{u}_{ro}) - \text{Im}(\mathbf{e}_s) \odot \text{Im}(\mathbf{u}_{ro}),   & \\
    \text{Im}(\mathbf{t}_\tau) \approx \text{Re}(\mathbf{e}_s) \odot \text{Im}(\mathbf{u}_{ro}) + \text{Im}(\mathbf{e}_s) \odot \text{Re}(\mathbf{u}_{ro}).
\end{split}
\end{equation}
We follow the same computations to infer the real and imaginary part of the desired (soft-supervised) time embeddings. Here, we treat $\mathbf{q}_{\text{time}}$ as a relation embedding $\mathbf{v}_r$ and the annotated entities as subject $s$ and  object $o$ interchangeably to generate $\mathbf{t}_1$ and $\mathbf{t}_2$, respectively. If either $s$ or $o$ is not present, we use dummy ones. We term this method \methods. 

Note here that the difference of hard and soft supervision relies on the available facts given during QA, i.e., access to the TKG. Moreover, soft-supervision may generalize better since it infers the temporal information in the embedding space. In Section~(\ref{sec:results}), we demonstrate the benefits and limitations of each approach.

\noindent
\textbf{Fusing temporal information.} \label{sec:fusion}
After obtaining time embeddings $\mathbf{t}_1$ and $\mathbf{t}_2$, we leverage them to enhance the question representation with temporal information. Specifically, we compute the $i$th collumn of the time-aware token embedding matrix $\mathbf{Q}_{T}$ as 
\begin{equation}
    \mathbf{q}_{T_i} = 
    \begin{cases}
    \mathbf{q}_{E_i}, & \text{if token $i$ is not an entity,}\\
      \mathbf{q}_{E_i} + \mathbf{t}_1 + \mathbf{t}_2, & \text{if token $i$ is an entity.} \\
\end{cases}
\label{eq:tempfusion}
\end{equation}
with $\mathbf{Q}_T:=[\mathbf{q}_{T_{\text{CLS}}}, \mathbf{q}_{T_1}, \ldots, \mathbf{q}_{T_N}]$.
Our intuition for summing the entity and time embeddings together follows the motivation of how transformer-based LMs, e.g., \textsc{BERT} use positional embedding for tokens ~\cite{vaswani2017transformer}. Here, time embeddings can be seen as entity positions in the time dimension. $\mathbf{Q}_{T}$ contains text, entity and time-aware information. Next, we propose an information fusion layer to combine this information altogether into a single question representation $\mathbf{q}$.


\subsection{Answer Prediction}
 Following~\cite{Fvry2020EntitiesAE}, we use an information fusion layer that consists of a dedicated learnable encoder $f(\cdot)$ which consists of $l$ Transformer encoding layers~\cite{vaswani2017transformer}. This encoder allows the question's tokens to attend to each other, which fuses context, entity and time-aware information together. The final token embedding matrix $\mathbf{Q}$ is calculated as 
 \begin{equation}
    \mathbf{Q} = f(\mathbf{Q}_{T}),
    \label{eq:Q3}
\end{equation}
where the columns of the embedding matrix correspond to the initial tokens $\mathbf{Q}:=[\mathbf{q}_{{\text{CLS}}}, \mathbf{q}_{1}, \ldots, \mathbf{q}_{N}]$.
As a final question representation, we use the embedding of the [CLS] token $\mathbf{q}:=\mathbf{q}_{{\text{CLS}}}$. 



The final score of an entity $\epsilon \in \mathcal{E}$ being the answer is given by 
\begin{align}
\begin{split}
    \max \;  \big( &\phi(\mathbf{e}_s,  \mathbf{P}_E\mathbf{q} , \mathbf{e}_\epsilon, \mathbf{t}_{\tau}), \\
    & \phi(\mathbf{e}_o, \mathbf{P}_E \mathbf{q}  , \mathbf{e}_\epsilon, \mathbf{t}_{\tau})\big),
\end{split}
\label{eq:phie}
\end{align}
where $s,o$ and $\tau$ are the annotated subject, object and timestamp, respectively, and  $\mathbf{P}_E $ is a $\ndims\times\ndims$ learnable matrix specific for entity predictions. Here, we treat the annotated subject and object interchangeably, and the $\max(\cdot)$ function ensures that we ignore the scores when $s$ or $o$ is a dummy entity.  

In addition, the final score of an timestamp $\tau \in \mathcal{T}$ being the answer is given by 
\begin{equation}
      \phi(\mathbf{e}_s, \mathbf{P}_T \mathbf{q}, \mathbf{e}_o, \mathbf{t}_{\tau}),
    \label{eq:phit}
\end{equation}
where $s,o$ are annotated entities in the question and $\mathbf{P}_T$ is a $\ndims\times\ndims$ learnable matrix specific for time predictions.
During training, the entity and time scores are concatenated and transformed to probabilities by a softmax function. The model's parameters are updated to assign higher probabilities to the correct answers by minimizing a cross entropy loss.

\input{tables/main_results}

\section{Experimental Setting}\label{sec:exps}

\textbf{Datasets}.
CronQuestions~\cite{saxena2021cronkgqa} is a temporal QA benchmark based on the Wikidata TKG proposed in~\cite{Lacroix2020Tcomplex}.
 The WikiData TKG consists of 125k entities, 203 relations, 1.7k timestamps (timestamps correspond to years), and 328k facts. In this TKG, facts are represented as (subject, relation, object, [start\_time, end\_time]). CronQuestions consists of 410k unique question-answer pairs, 350k of which are for training and 30k for validation and for testing. Moreover, the entities and times present in the questions are annotated. CronQuestions includes both simple and  complex temporal questions ( Table~\ref{tab:cronqs} for examples).

\noindent
\textbf{Incomplete WikiData TKG.}
To illustrate how methods perform under incomplete TKGs, we provide a setting where the given WikiData TKG is corrupted at the time dimension. Specifically, for a fact (subject, relation, object, [start\_time, end\_time]) $\in \mathcal{F}$, we remove the associated timestamps with probability $p$. If the timestamps are removed, the fact becomes (subject, relation, object, no\_time). Here, `no\_time' denotes that there is no timestamp associated with (subject, relation, object) and we treat it as a special timestamp. 

We used the corrupted TKG to perform two experiments. In the first experiment, we substitute the original TKG with the corrupted one during the QA task. This affects only \method-Hard since this is the only method that uses a TKG during QA. The second configuration is to substitute the original TKG with the corrupted one throughout the process. This means that TComplEx embeddings are generated on a corrupted TKG and, thus, may not encode important temporal information. All TKGQA embedding-based methods are affected by this configuration.

\noindent
\textbf{Additional Complex Questions.} Although CronQuestions includes different question types, we manually create additional complex types. The idea is to evaluate how different methods perform on complex questions that were unseen during training but include the same keywords (and temporal constraints) with the training questions. We create (i) `\textit{before \& after}' questions that include both `before' and `after' constraints and (ii) `\textit{before/after \& first/last}' questions that include both `before/after' and `first/last' constraints. We describe the details of generating these QA pairs in the Appendix.



\subsection{Model Configuration}
We learn TKG embeddings with the TComplEx method, where we set their dimensions $\ndims=512$. During, QA the pre-trained LM's parameters and the TKG embeddings are not updated. We set the number of transformer layers of the encoder $f(\cdot)$ to $l=6$ with 8 heads per layer. We also observed the same performance when setting $l=3$ with 4 heads per layer. The model's parameters are  updated with Adam~\cite{kingma2014adam} with a learning rate of 0.0002. The model is trained for 20 maximum epochs and the final parameters are determined based on the best validation performance. The model is implemented with Pytorch~\cite{Pytorch}. For reproducibility, our code is available at: \url{https://github.com/cmavro/TempoQR}.


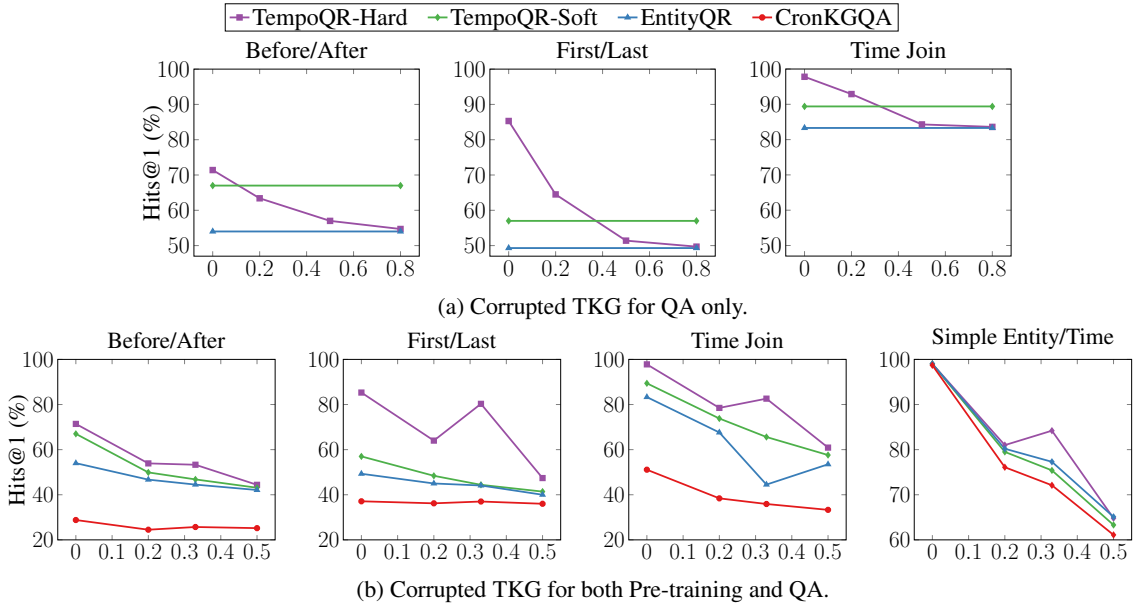
\begin{figure*}
  \centering
  \begin{subfigure}[t]{0.7\textwidth}
    \resizebox{.95\linewidth}{!}{\input{figures/random_times.tex}}
    \caption{Corrupted TKG for QA only.}
    \label{fig:random}
    \end{subfigure}
    \hfill
    \begin{subfigure}[t]{0.9\textwidth}
        \resizebox{.95\linewidth}{!}{\input{figures/missing_link.tex}}
        \caption{Corrupted TKG for both Pre-training and QA.}
        \label{fig:missing}
    \end{subfigure}
    \caption{Performance comparison when the underlying TKG is corrupted. x-axis corresponds to the probability $p$ of a fact to be corrupted.}
\end{figure*}


\subsection{Baseline methods and \method{} variations}

\textbf{Pre-trained LMs}. BERT~\cite{devlin2019bert} and RoBERTa~\cite{Liu2019RoBERTaAR} are two well-established pre-trained LMs. To evaluate these models, we generate their LM-based question embedding and concatenate it with the annotated entity and time embeddings, followed by a learnable projection. The resulted embedding is scored against all entities and timestamps via dot-product.
    
\noindent\textbf{EaE and \baseline}.  Entity as Experts (EaE)~\cite{Fvry2020EntitiesAE} is an entity-aware method similar to \method{}. The key differences are that EaE does not utilize a TKG embedding-based scoring function for answer prediction and that it does not fuse additional temporal information as in Section~\ref{sec:TAQE}. As a baseline, we experiment with EaE combined with TComplEx scoring function. Since this baseline is  similar to \method{} without the steps of Section~(\ref{sec:TAQE}), we call it \baseline.
    
\noindent\textbf{CronKGQA and EmbedKGQA}. CronKGQA is the TKGQA embedding-based method described in Section~\ref{sec:tkgqa}. EmbedKGQA~\cite{saxena2020embedkgqa} is similar to CronKGQA, but designed for regular KGs. In~\cite{saxena2021cronkgqa}, EmbedKGQA is implemented for TKGQA as follows. Timestamps are ignored during pre-training and random time embeddings are used during the QA task.

\noindent\textbf{CronKGQA and \baseline with hard and soft supervision}. CronKGQA and \baseline{} are extended to incorporate additional temporal information by the algorithmic steps in Section~\ref{sec:TAQE}, which generate time embeddings $\mathbf{t}_1$ and $\mathbf{t}_2$ for \method. Recall that \method generates $\mathbf{t}_1$ and $\mathbf{t}_2$ by either accessing the TKG or by inferring them in the embedding space. For CronKGQA and \baseline, we generate $\mathbf{t}_1$ and $\mathbf{t}_2$ in the same way, but we employ them directly to the TComplEx scoring function as follows. We modify~\eqref{eq:phie}, which scores an entity to be the answer, as
    \begin{align}
    \begin{split}
        \max \;  \big( &\phi(\mathbf{e}_s,  \mathbf{P}_E\mathbf{q} , \mathbf{e}_\epsilon, \mathbf{t}_1 + \mathbf{t}_2), \\
        & \phi(\mathbf{e}_o, \mathbf{P}_E \mathbf{q}  , \mathbf{e}_\epsilon, \mathbf{t}_1+\mathbf{t}_2)\big),
    \end{split}
\end{align}
    to replace the embedding of a dummy timestamp $\mathbf{t}_\tau$ (if no time is annotated in the question) with $\mathbf{t}_1 + \mathbf{t}_2$. Based on how $\mathbf{t}_1$ and $\mathbf{t}_2$ are generated (hard or soft supervision), we term the methods CronKGQA-Hard and \baseline-Hard or CronKGQA-Soft and \baseline-Soft, respectively.

\section{Results}\label{sec:results}
\subsection{Main Results}

 Table~\ref{tab:main-results} shows the results of our method compared to other baselines on CronQuestions. First, by comparing \baseline{} to CronKGQA, we see that grounding the question to the entities it refers (entity-aware step) significantly helps for answering complex questions. In this case, the absolute improvement for complex questions is 17\% and 10\% at Hits@1 and Hits@10, respectively. Furthermore, comparing \method{} to \baseline{}, we see the benefit of adding temporal information to the question (time-aware step). The absolute improvement of \methods{} over \baseline{} is 9\% at Hits@1 for complex questions, while the respective improvement of \methodh{}  is more than 30\%. Moreover, \methodh{} outperforms \methods{} by 25\% at Hits@1 when the answer is a time. This confirms that \methodh{} provides accurate temporal information by retrieving it from the TKG, while \methods{} sometimes cannot infer as accurate information from the embedding space.

We also highlight that methods that score possible answers with the TComplEx function (\method{}, \baseline{} and CronKGQA) answer  99\% of the simple questions correctly. The other methods (BERT, RoBERTa, EmbedKGQA and EaE) cannot answer correctly more the 35\% (Hits@1) and 76\% (Hits@10). Similarly, BERT, RoBERTa, EmbedKGQA and EaE have 35\%-65\% and 11\%-40\% worse overall accuracy for Hits@1 and Hits@10, respectively, compared to \method{}, \baseline{} and CronKGQA.

\input{tables/hard_soft}
\begin{figure}
\centering
\resizebox{\columnwidth}{!}{\input{figures/before_after.tex}}
\caption{Evaluation for unseen complex types during inference. x-axis corresponds to $k$ of Hits@$k$. }
\label{fig:bef_aft}
\end{figure}
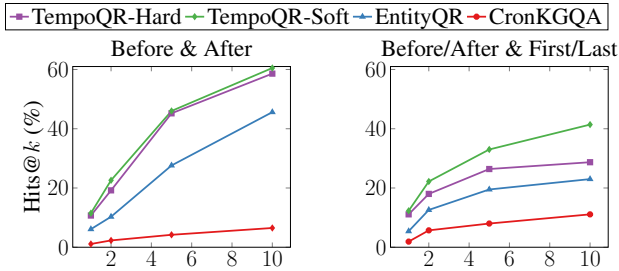
\subsection{Ablation Study}


Table~\ref{tab:hard-soft} shows how soft and hard supervision affect the performance of various methods for different question types. First, we see that both supervision approaches have a positive effect on CronKGQA, where the performance is improved by 2.5\% and 6\% over the original method (CronKGQA-Soft and CronKGQA-Hard, respectively, compared to CronKGQA). The same does not happen for \baseline{}, where its performance drops (\baseline-Soft and \baseline-Hard  compared to \baseline). This is an effect of over-using TKG embeddings, since \baseline{}-Soft and \baseline{}-Hard use this information multiple times for answer prediction.

Moreover, \methodh{} performs much better for `first/last' questions compared to soft-supervision (27\% absolute improvement). Since \methodh{}  always retrieves the start and end timestamps of the question entities, this provides more accurate temporal information to \textit{``When was the first/last time..."} questions. On the other hand, this does not equally benefit `before/after' questions;  the improvement of hard-supervision over soft-supervision is only 3\%. This indicates that both approaches handle such questions in a similar manner. Finally, \method-Soft performs 8\% worse in `time join' questions compared to \method-Hard. This indicates that might be some room for improvement for inferring more accurate time embeddings with soft-supervision.

\subsection{Robustness over Corrupted TKGs}

Figure~\ref{fig:random} shows the results when the given TKG is corrupted during the QA phase. \method-Hard is the only method that depends on the quality of the TKG during QA and is greatly affected by a corrupted TKG. When a lot of facts are corrupted ($p=0.8$) it performs similar to \baseline{}, which does not use any additional temporal information. For `before/after' questions, it performs worse than \methods even when the facts are corrupted by a probability $p=0.2$. Finally, \methodh is more robust for `first/last' and `time join' questions, where it better handles the non-corrupted timestamps of the facts.

Figure~\ref{fig:missing} shows the results when a corrupted TKG is given for both TKG embedding and QA. We can see that methods that rely on TKG embeddings (\methodh, \methods) are greatly affected, since they perform similar to each other as well to \baseline{} when the corruption probability $p$ is large, e.g., $p=0.5$. CornKGQA is the least affected by a corrupted TKG, but, still, its performance is much lower. The largest performance drop is observed for \methodh{} for `first/last' questions, which indicates that it cannot generalize well to such questions under a corrupted TKG. Finally, depending on the information that is corrupted, some methods benefit over others, i.e., compare \methodh{} to \baseline{} for $p=0.33$ at `time join' questions.

\subsection{Unseen Question Types}
Figure~\ref{fig:bef_aft} shows the performance for unseen question types during training. As we can see, \methods{} performs the best for both generated question types. Since it learns to handle temporal constraints in the embedding space, it generalizes better than \methodh{} which simply uses the TKG to answer such questions. In general, neither of these methods  seems to be able to tackle unseen questions effectively, i.e., Hits@1 is below 20\% for all the methods. This is also confirmed since the performance of the methods only increases when $k$ is increased, i.e., the return more possible answers. This motivates the need for extending these methods in a way that ensures better performance for unseen questions. 

\section{Conclusion}
This paper puts forth a comprehensive embedding framework specialized in answering complex questions over TKGs. The benefit of \method comes by learning context, entity and time-aware question representations. The latter relies either on hard or soft supervision. Extensive experiments confirmed the benefits of each step performed in our method. \method outperforms existing methods for complex questions by 25--45\%. The limitations and advantages of hard and soft supervision are also showcased. Future research includes extending existing methods to generalize better to unseen question types.  



\bibliography{aaai22}

\input{appendix}

\end{document}

%% file: define.tex
\newcommand{\method}{TempoQR\xspace}
\newcommand{\methods}{TempoQR-Soft\xspace}
\newcommand{\methodh}{TempoQR-Hard\xspace}

\newcommand{\baseline}{EntityQR\xspace}

\newcommand{\ntokens}{N}
\newcommand{\ndims}{D}
\newcommand{\bndims}{D_B}


%% file: figures/framework/phase1.tex
\tikzset{every picture/.style={line width=0.75pt}} 

\begin{tikzpicture}[x=0.7pt,y=0.3pt,yscale=-1,xscale=1]

\draw    (58,384) -- (58.99,10) ;
\draw [shift={(59,8)}, rotate = 450.15] [color={rgb, 255:red, 0; green, 0; blue, 0 }  ][line width=0.75]    (10.93,-3.29) .. controls (6.95,-1.4) and (3.31,-0.3) .. (0,0) .. controls (3.31,0.3) and (6.95,1.4) .. (10.93,3.29)   ;
\draw    (58,384) -- (417,383.01) ;
\draw [shift={(419,383)}, rotate = 539.8399999999999] [color={rgb, 255:red, 0; green, 0; blue, 0 }  ][line width=0.75]    (10.93,-3.29) .. controls (6.95,-1.4) and (3.31,-0.3) .. (0,0) .. controls (3.31,0.3) and (6.95,1.4) .. (10.93,3.29)   ;
\draw  [fill={rgb, 255:red, 245; green, 166; blue, 35 }  ,fill opacity=1 ] (106,188) .. controls (106,183.58) and (109.58,180) .. (114,180) .. controls (118.42,180) and (122,183.58) .. (122,188) .. controls (122,192.42) and (118.42,196) .. (114,196) .. controls (109.58,196) and (106,192.42) .. (106,188) -- cycle ;
\draw  [fill={rgb, 255:red, 208; green, 2; blue, 27 }  ,fill opacity=1 ] (113.63,251.7) .. controls (113.63,247.18) and (117.3,243.52) .. (121.82,243.52) .. controls (126.34,243.52) and (130,247.18) .. (130,251.7) .. controls (130,256.22) and (126.34,259.88) .. (121.82,259.88) .. controls (117.3,259.88) and (113.63,256.22) .. (113.63,251.7) -- cycle ;
\draw  [fill={rgb, 255:red, 144; green, 19; blue, 254 }  ,fill opacity=1 ] (303,242) .. controls (303,237.58) and (306.58,234) .. (311,234) .. controls (315.42,234) and (319,237.58) .. (319,242) .. controls (319,246.42) and (315.42,250) .. (311,250) .. controls (306.58,250) and (303,246.42) .. (303,242) -- cycle ;
\draw  [fill={rgb, 255:red, 74; green, 144; blue, 226 }  ,fill opacity=1 ] (251,143) .. controls (251,138.58) and (254.58,135) .. (259,135) .. controls (263.42,135) and (267,138.58) .. (267,143) .. controls (267,147.42) and (263.42,151) .. (259,151) .. controls (254.58,151) and (251,147.42) .. (251,143) -- cycle ;
\draw  [fill={rgb, 255:red, 184; green, 233; blue, 134 }  ,fill opacity=1 ] (145,67) .. controls (145,62.58) and (148.58,59) .. (153,59) .. controls (157.42,59) and (161,62.58) .. (161,67) .. controls (161,71.42) and (157.42,75) .. (153,75) .. controls (148.58,75) and (145,71.42) .. (145,67) -- cycle ;
\draw  [fill={rgb, 255:red, 80; green, 227; blue, 194 }  ,fill opacity=1 ] (262,64) .. controls (262,59.58) and (265.58,56) .. (270,56) .. controls (274.42,56) and (278,59.58) .. (278,64) .. controls (278,68.42) and (274.42,72) .. (270,72) .. controls (265.58,72) and (262,68.42) .. (262,64) -- cycle ;
\draw  [fill={rgb, 255:red, 248; green, 231; blue, 28 }  ,fill opacity=1 ] (318,319) .. controls (318,314.58) and (321.58,311) .. (326,311) .. controls (330.42,311) and (334,314.58) .. (334,319) .. controls (334,323.42) and (330.42,327) .. (326,327) .. controls (321.58,327) and (318,323.42) .. (318,319) -- cycle ;
\draw    (119,183) -- (152.4,76.91) ;
\draw [shift={(153,75)}, rotate = 467.47] [color={rgb, 255:red, 0; green, 0; blue, 0 }  ][line width=0.75]    (10.93,-3.29) .. controls (6.95,-1.4) and (3.31,-0.3) .. (0,0) .. controls (3.31,0.3) and (6.95,1.4) .. (10.93,3.29)   ;
\draw    (119,183) -- (247.07,148.52) ;
\draw [shift={(249,148)}, rotate = 524.9300000000001] [color={rgb, 255:red, 0; green, 0; blue, 0 }  ][line width=0.75]    (10.93,-3.29) .. controls (6.95,-1.4) and (3.31,-0.3) .. (0,0) .. controls (3.31,0.3) and (6.95,1.4) .. (10.93,3.29)   ;
\draw    (114,196) -- (120.67,236.03) ;
\draw [shift={(121,238)}, rotate = 260.54] [color={rgb, 255:red, 0; green, 0; blue, 0 }  ][line width=0.75]    (10.93,-3.29) .. controls (6.95,-1.4) and (3.31,-0.3) .. (0,0) .. controls (3.31,0.3) and (6.95,1.4) .. (10.93,3.29)   ;
\draw    (130,251.7) -- (301,242.11) ;
\draw [shift={(303,242)}, rotate = 536.79] [color={rgb, 255:red, 0; green, 0; blue, 0 }  ][line width=0.75]    (10.93,-3.29) .. controls (6.95,-1.4) and (3.31,-0.3) .. (0,0) .. controls (3.31,0.3) and (6.95,1.4) .. (10.93,3.29)   ;
\draw    (161,67) -- (253,66.02) ;
\draw [shift={(255,66)}, rotate = 539.39] [color={rgb, 255:red, 0; green, 0; blue, 0 }  ][line width=0.75]    (10.93,-3.29) .. controls (6.95,-1.4) and (3.31,-0.3) .. (0,0) .. controls (3.31,0.3) and (6.95,1.4) .. (10.93,3.29)   ;
\draw    (326,311) -- (314.42,256.96) ;
\draw [shift={(314,255)}, rotate = 437.91] [color={rgb, 255:red, 0; green, 0; blue, 0 }  ][line width=0.75]    (10.93,-3.29) .. controls (6.95,-1.4) and (3.31,-0.3) .. (0,0) .. controls (3.31,0.3) and (6.95,1.4) .. (10.93,3.29)   ;

\draw (86,265) node [anchor=north west][inner sep=0.75pt]   [align=left] {`The Godfather'};
\draw (60,140) node [anchor=north west][inner sep=0.75pt]   [align=left] {Best \\Picture};
\draw (298,200) node [anchor=north west][inner sep=0.75pt]   [align=left] {Marlon Brando};
\draw (259,114) node [anchor=north west][inner sep=0.75pt]   [align=left] {`The Sting'};
\draw (119,25) node [anchor=north west][inner sep=0.75pt]   [align=left] {`Forrest Gump'};
\draw (280,42) node [anchor=north west][inner sep=0.75pt]   [align=left] {Tom Hanks};
\draw (286,330) node [anchor=north west][inner sep=0.75pt]   [align=left] {`Apocalypse Now'};
\draw (125,200) node [anchor=north west][inner sep=0.75pt]   [align=left] {\textit{{\fontfamily{ptm}\selectfont won,1972}}};
\draw (80,90) node [anchor=north west][inner sep=0.75pt]   [align=left] {\textit{{\fontfamily{ptm}\selectfont won,1994}}};
\draw (163,125) node [anchor=north west][inner sep=0.75pt]   [align=left] {\textit{{\fontfamily{ptm}\selectfont won,1973}}};
\draw (210,245) node [anchor=north west][inner sep=0.75pt]   [align=left] {\textit{{\fontfamily{ptm}\selectfont actor,1972}}};
\draw (325,276) node [anchor=north west][inner sep=0.75pt]   [align=left] {\textit{{\fontfamily{ptm}\selectfont actor,1979}}};
\draw (176,65) node [anchor=north west][inner sep=0.75pt]   [align=left] {\textit{{\fontfamily{ptm}\selectfont actor,1994}}};

\end{tikzpicture}

%% file: figures/framework/phase2.tex
\tikzset{every picture/.style={line width=0.75pt}} 

\begin{tikzpicture}[x=0.7pt,y=0.3pt,yscale=-1,xscale=1]

\draw    (58,384) -- (58.99,10) ;
\draw [shift={(59,8)}, rotate = 450.15] [color={rgb, 255:red, 0; green, 0; blue, 0 }  ][line width=0.75]    (10.93,-3.29) .. controls (6.95,-1.4) and (3.31,-0.3) .. (0,0) .. controls (3.31,0.3) and (6.95,1.4) .. (10.93,3.29)   ;
\draw  [fill={rgb, 255:red, 245; green, 166; blue, 35 }  ,fill opacity=1 ] (106,188) .. controls (106,183.58) and (109.58,180) .. (114,180) .. controls (118.42,180) and (122,183.58) .. (122,188) .. controls (122,192.42) and (118.42,196) .. (114,196) .. controls (109.58,196) and (106,192.42) .. (106,188) -- cycle ;
\draw  [fill={rgb, 255:red, 208; green, 2; blue, 27 }  ,fill opacity=1 ] (113.63,251.7) .. controls (113.63,247.18) and (117.3,243.52) .. (121.82,243.52) .. controls (126.34,243.52) and (130,247.18) .. (130,251.7) .. controls (130,256.22) and (126.34,259.88) .. (121.82,259.88) .. controls (117.3,259.88) and (113.63,256.22) .. (113.63,251.7) -- cycle ;
\draw  [fill={rgb, 255:red, 144; green, 19; blue, 254 }  ,fill opacity=1 ] (303,242) .. controls (303,237.58) and (306.58,234) .. (311,234) .. controls (315.42,234) and (319,237.58) .. (319,242) .. controls (319,246.42) and (315.42,250) .. (311,250) .. controls (306.58,250) and (303,246.42) .. (303,242) -- cycle ;
\draw  [fill={rgb, 255:red, 74; green, 144; blue, 226 }  ,fill opacity=1 ] (251,143) .. controls (251,138.58) and (254.58,135) .. (259,135) .. controls (263.42,135) and (267,138.58) .. (267,143) .. controls (267,147.42) and (263.42,151) .. (259,151) .. controls (254.58,151) and (251,147.42) .. (251,143) -- cycle ;
\draw  [fill={rgb, 255:red, 184; green, 233; blue, 134 }  ,fill opacity=1 ] (145,67) .. controls (145,62.58) and (148.58,59) .. (153,59) .. controls (157.42,59) and (161,62.58) .. (161,67) .. controls (161,71.42) and (157.42,75) .. (153,75) .. controls (148.58,75) and (145,71.42) .. (145,67) -- cycle ;
\draw  [fill={rgb, 255:red, 80; green, 227; blue, 194 }  ,fill opacity=1 ] (262,64) .. controls (262,59.58) and (265.58,56) .. (270,56) .. controls (274.42,56) and (278,59.58) .. (278,64) .. controls (278,68.42) and (274.42,72) .. (270,72) .. controls (265.58,72) and (262,68.42) .. (262,64) -- cycle ;
\draw  [fill={rgb, 255:red, 248; green, 231; blue, 28 }  ,fill opacity=1 ] (318,319) .. controls (318,314.58) and (321.58,311) .. (326,311) .. controls (330.42,311) and (334,314.58) .. (334,319) .. controls (334,323.42) and (330.42,327) .. (326,327) .. controls (321.58,327) and (318,323.42) .. (318,319) -- cycle ;
\draw  [fill={rgb, 255:red, 139; green, 87; blue, 42 }  ,fill opacity=0.28 ] (91.66,209.38) .. controls (91.7,183.54) and (103.49,167.37) .. (118.01,173.26) .. controls (132.53,179.15) and (144.27,204.87) .. (144.24,230.71) .. controls (144.21,256.55) and (132.42,272.72) .. (117.9,266.84) .. controls (103.38,260.95) and (91.63,235.22) .. (91.66,209.38) -- cycle ;
\draw  [dash pattern={on 4.5pt off 4.5pt}]  (130,177) .. controls (169.6,147.3) and (109.23,108.78) .. (146.83,78.9) ;
\draw [shift={(148,78)}, rotate = 503.13] [color={rgb, 255:red, 0; green, 0; blue, 0 }  ][line width=0.75]    (10.93,-3.29) .. controls (6.95,-1.4) and (3.31,-0.3) .. (0,0) .. controls (3.31,0.3) and (6.95,1.4) .. (10.93,3.29)   ;
\draw  [dash pattern={on 4.5pt off 4.5pt}]  (136,183) .. controls (175.6,153.3) and (205.4,96.16) .. (254.51,71.73) ;
\draw [shift={(256,71)}, rotate = 514.36] [color={rgb, 255:red, 0; green, 0; blue, 0 }  ][line width=0.75]    (10.93,-3.29) .. controls (6.95,-1.4) and (3.31,-0.3) .. (0,0) .. controls (3.31,0.3) and (6.95,1.4) .. (10.93,3.29)   ;
\draw  [dash pattern={on 4.5pt off 4.5pt}]  (141,198) .. controls (180.6,168.3) and (204.52,185.64) .. (253.51,162.71) ;
\draw [shift={(255,162)}, rotate = 514.36] [color={rgb, 255:red, 0; green, 0; blue, 0 }  ][line width=0.75]    (10.93,-3.29) .. controls (6.95,-1.4) and (3.31,-0.3) .. (0,0) .. controls (3.31,0.3) and (6.95,1.4) .. (10.93,3.29)   ;
\draw  [dash pattern={on 4.5pt off 4.5pt}]  (143,206) .. controls (182.6,176.3) and (240.82,261.27) .. (290.5,239.69) ;
\draw [shift={(292,239)}, rotate = 514.36] [color={rgb, 255:red, 0; green, 0; blue, 0 }  ][line width=0.75]    (10.93,-3.29) .. controls (6.95,-1.4) and (3.31,-0.3) .. (0,0) .. controls (3.31,0.3) and (6.95,1.4) .. (10.93,3.29)   ;
\draw  [dash pattern={on 4.5pt off 4.5pt}]  (145,222) .. controls (184.6,192.3) and (250.66,321.38) .. (304.38,315.23) ;
\draw [shift={(306,315)}, rotate = 530.54] [color={rgb, 255:red, 0; green, 0; blue, 0 }  ][line width=0.75]    (10.93,-3.29) .. controls (6.95,-1.4) and (3.31,-0.3) .. (0,0) .. controls (3.31,0.3) and (6.95,1.4) .. (10.93,3.29)   ;
\draw    (58,384) -- (417,383.01) ;
\draw [shift={(419,383)}, rotate = 539.8399999999999] [color={rgb, 255:red, 0; green, 0; blue, 0 }  ][line width=0.75]    (10.93,-3.29) .. controls (6.95,-1.4) and (3.31,-0.3) .. (0,0) .. controls (3.31,0.3) and (6.95,1.4) .. (10.93,3.29)   ;

\draw (86,280) node [anchor=north west][inner sep=0.75pt]   [align=left] {`The Godfather'};
\draw (60,110) node [anchor=north west][inner sep=0.75pt]   [align=left] {Best \\Picture};
\draw (298,200) node [anchor=north west][inner sep=0.75pt]   [align=left] {Marlon Brando};
\draw (259,114) node [anchor=north west][inner sep=0.75pt]   [align=left] {`The Sting'};
\draw (119,25) node [anchor=north west][inner sep=0.75pt]   [align=left] {`Forrest Gump'};
\draw (280,42) node [anchor=north west][inner sep=0.75pt]   [align=left] {Tom Hanks};
\draw (286,330) node [anchor=north west][inner sep=0.75pt]   [align=left] {`Apocalypse Now'};
\draw (103,79) node [anchor=north west][inner sep=0.75pt]   [align=left] {(0.2)};
\draw (202,135) node [anchor=north west][inner sep=0.75pt]   [align=left] {(0.3)};
\draw (230,300) node [anchor=north west][inner sep=0.75pt]   [align=left] {(0.25)};
\draw (176,81) node [anchor=north west][inner sep=0.75pt]   [align=left] {(0.1)};
\draw (233,200) node [anchor=north west][inner sep=0.75pt]   [align=left] {(0.15)};
\end{tikzpicture}

%% file: figures/framework/phase3.tex
\tikzset{every picture/.style={line width=0.75pt}} 

\begin{tikzpicture}[x=0.7pt,y=0.3pt,yscale=-1,xscale=1]

\draw    (59,397) -- (59.99,23) ;
\draw [shift={(60,21)}, rotate = 450.15] [color={rgb, 255:red, 0; green, 0; blue, 0 }  ][line width=0.75]    (10.93,-3.29) .. controls (6.95,-1.4) and (3.31,-0.3) .. (0,0) .. controls (3.31,0.3) and (6.95,1.4) .. (10.93,3.29)   ;
\draw  [fill={rgb, 255:red, 245; green, 166; blue, 35 }  ,fill opacity=1 ] (107,201) .. controls (107,196.58) and (110.58,193) .. (115,193) .. controls (119.42,193) and (123,196.58) .. (123,201) .. controls (123,205.42) and (119.42,209) .. (115,209) .. controls (110.58,209) and (107,205.42) .. (107,201) -- cycle ;
\draw  [fill={rgb, 255:red, 208; green, 2; blue, 27 }  ,fill opacity=1 ] (114.63,264.7) .. controls (114.63,260.18) and (118.3,256.52) .. (122.82,256.52) .. controls (127.34,256.52) and (131,260.18) .. (131,264.7) .. controls (131,269.22) and (127.34,272.88) .. (122.82,272.88) .. controls (118.3,272.88) and (114.63,269.22) .. (114.63,264.7) -- cycle ;
\draw  [fill={rgb, 255:red, 144; green, 19; blue, 254 }  ,fill opacity=1 ] (304,255) .. controls (304,250.58) and (307.58,247) .. (312,247) .. controls (316.42,247) and (320,250.58) .. (320,255) .. controls (320,259.42) and (316.42,263) .. (312,263) .. controls (307.58,263) and (304,259.42) .. (304,255) -- cycle ;
\draw  [fill={rgb, 255:red, 74; green, 144; blue, 226 }  ,fill opacity=1 ] (252,156) .. controls (252,151.58) and (255.58,148) .. (260,148) .. controls (264.42,148) and (268,151.58) .. (268,156) .. controls (268,160.42) and (264.42,164) .. (260,164) .. controls (255.58,164) and (252,160.42) .. (252,156) -- cycle ;
\draw  [fill={rgb, 255:red, 184; green, 233; blue, 134 }  ,fill opacity=1 ] (146,80) .. controls (146,75.58) and (149.58,72) .. (154,72) .. controls (158.42,72) and (162,75.58) .. (162,80) .. controls (162,84.42) and (158.42,88) .. (154,88) .. controls (149.58,88) and (146,84.42) .. (146,80) -- cycle ;
\draw  [fill={rgb, 255:red, 80; green, 227; blue, 194 }  ,fill opacity=1 ] (263,77) .. controls (263,72.58) and (266.58,69) .. (271,69) .. controls (275.42,69) and (279,72.58) .. (279,77) .. controls (279,81.42) and (275.42,85) .. (271,85) .. controls (266.58,85) and (263,81.42) .. (263,77) -- cycle ;
\draw  [fill={rgb, 255:red, 248; green, 231; blue, 28 }  ,fill opacity=1 ] (319,332) .. controls (319,327.58) and (322.58,324) .. (327,324) .. controls (331.42,324) and (335,327.58) .. (335,332) .. controls (335,336.42) and (331.42,340) .. (327,340) .. controls (322.58,340) and (319,336.42) .. (319,332) -- cycle ;
\draw  [fill={rgb, 255:red, 139; green, 87; blue, 42 }  ,fill opacity=0.28 ] (92.66,222.38) .. controls (92.7,196.54) and (104.49,180.37) .. (119.01,186.26) .. controls (133.53,192.15) and (145.27,217.87) .. (145.24,243.71) .. controls (145.21,269.55) and (133.42,285.72) .. (118.9,279.84) .. controls (104.38,273.95) and (92.63,248.22) .. (92.66,222.38) -- cycle ;
\draw  [dash pattern={on 4.5pt off 4.5pt}]  (131,190) .. controls (170.6,160.3) and (110.23,121.78) .. (147.83,91.9) ;
\draw [shift={(149,91)}, rotate = 503.13] [color={rgb, 255:red, 0; green, 0; blue, 0 }  ][line width=0.75]    (10.93,-3.29) .. controls (6.95,-1.4) and (3.31,-0.3) .. (0,0) .. controls (3.31,0.3) and (6.95,1.4) .. (10.93,3.29)   ;
\draw  [dash pattern={on 4.5pt off 4.5pt}]  (142,211) .. controls (181.6,181.3) and (205.52,198.64) .. (254.51,175.71) ;
\draw [shift={(256,175)}, rotate = 514.36] [color={rgb, 255:red, 0; green, 0; blue, 0 }  ][line width=0.75]    (10.93,-3.29) .. controls (6.95,-1.4) and (3.31,-0.3) .. (0,0) .. controls (3.31,0.3) and (6.95,1.4) .. (10.93,3.29)   ;
\draw  [dash pattern={on 4.5pt off 4.5pt}]  (146,235) .. controls (185.6,205.3) and (251.66,334.38) .. (305.38,328.23) ;
\draw [shift={(307,328)}, rotate = 530.54] [color={rgb, 255:red, 0; green, 0; blue, 0 }  ][line width=0.75]    (10.93,-3.29) .. controls (6.95,-1.4) and (3.31,-0.3) .. (0,0) .. controls (3.31,0.3) and (6.95,1.4) .. (10.93,3.29)   ;
\draw    (59,397) -- (418,396.01) ;
\draw [shift={(420,396)}, rotate = 539.8399999999999] [color={rgb, 255:red, 0; green, 0; blue, 0 }  ][line width=0.75]    (10.93,-3.29) .. controls (6.95,-1.4) and (3.31,-0.3) .. (0,0) .. controls (3.31,0.3) and (6.95,1.4) .. (10.93,3.29)   ;

\draw (87,285) node [anchor=north west][inner sep=0.75pt]   [align=left] {`The Godfather'};
\draw (61,125) node [anchor=north west][inner sep=0.75pt]   [align=left] {Best \\Picture};
\draw (299,215) node [anchor=north west][inner sep=0.75pt]   [align=left] {Marlon Brando (0.05)};
\draw (260,127) node [anchor=north west][inner sep=0.75pt]   [align=left] {`The Sting'};
\draw (120,40) node [anchor=north west][inner sep=0.75pt]   [align=left] {`Forrest Gump'};
\draw (281,55) node [anchor=north west][inner sep=0.75pt]   [align=left] {Tom Hanks (0.05)};
\draw (287,343) node [anchor=north west][inner sep=0.75pt]   [align=left] {`Apocalypse Now'};
\draw (103,92) node [anchor=north west][inner sep=0.75pt]   [align=left] {(0.4)};
\draw (203,150) node [anchor=north west][inner sep=0.75pt]   [align=left] {(0.4)};
\draw (230,313) node [anchor=north west][inner sep=0.75pt]   [align=left] {(0.1)};

\end{tikzpicture}

%% file: figures/framework/phase4.tex
\tikzset{every picture/.style={line width=0.75pt}} 

\begin{tikzpicture}[x=0.7pt,y=0.3pt,yscale=-1,xscale=1]

\draw    (59,397) -- (59.99,23) ;
\draw [shift={(60,21)}, rotate = 450.15] [color={rgb, 255:red, 0; green, 0; blue, 0 }  ][line width=0.75]    (10.93,-3.29) .. controls (6.95,-1.4) and (3.31,-0.3) .. (0,0) .. controls (3.31,0.3) and (6.95,1.4) .. (10.93,3.29)   ;
\draw  [fill={rgb, 255:red, 245; green, 166; blue, 35 }  ,fill opacity=1 ] (107,201) .. controls (107,196.58) and (110.58,193) .. (115,193) .. controls (119.42,193) and (123,196.58) .. (123,201) .. controls (123,205.42) and (119.42,209) .. (115,209) .. controls (110.58,209) and (107,205.42) .. (107,201) -- cycle ;
\draw  [fill={rgb, 255:red, 208; green, 2; blue, 27 }  ,fill opacity=1 ] (114.63,264.7) .. controls (114.63,260.18) and (118.3,256.52) .. (122.82,256.52) .. controls (127.34,256.52) and (131,260.18) .. (131,264.7) .. controls (131,269.22) and (127.34,272.88) .. (122.82,272.88) .. controls (118.3,272.88) and (114.63,269.22) .. (114.63,264.7) -- cycle ;
\draw  [fill={rgb, 255:red, 144; green, 19; blue, 254 }  ,fill opacity=1 ] (304,255) .. controls (304,250.58) and (307.58,247) .. (312,247) .. controls (316.42,247) and (320,250.58) .. (320,255) .. controls (320,259.42) and (316.42,263) .. (312,263) .. controls (307.58,263) and (304,259.42) .. (304,255) -- cycle ;
\draw  [fill={rgb, 255:red, 74; green, 144; blue, 226 }  ,fill opacity=1 ] (252,156) .. controls (252,151.58) and (255.58,148) .. (260,148) .. controls (264.42,148) and (268,151.58) .. (268,156) .. controls (268,160.42) and (264.42,164) .. (260,164) .. controls (255.58,164) and (252,160.42) .. (252,156) -- cycle ;
\draw  [fill={rgb, 255:red, 184; green, 233; blue, 134 }  ,fill opacity=1 ] (146,80) .. controls (146,75.58) and (149.58,72) .. (154,72) .. controls (158.42,72) and (162,75.58) .. (162,80) .. controls (162,84.42) and (158.42,88) .. (154,88) .. controls (149.58,88) and (146,84.42) .. (146,80) -- cycle ;
\draw  [fill={rgb, 255:red, 80; green, 227; blue, 194 }  ,fill opacity=1 ] (263,77) .. controls (263,72.58) and (266.58,69) .. (271,69) .. controls (275.42,69) and (279,72.58) .. (279,77) .. controls (279,81.42) and (275.42,85) .. (271,85) .. controls (266.58,85) and (263,81.42) .. (263,77) -- cycle ;
\draw  [fill={rgb, 255:red, 248; green, 231; blue, 28 }  ,fill opacity=1 ] (319,332) .. controls (319,327.58) and (322.58,324) .. (327,324) .. controls (331.42,324) and (335,327.58) .. (335,332) .. controls (335,336.42) and (331.42,340) .. (327,340) .. controls (322.58,340) and (319,336.42) .. (319,332) -- cycle ;
\draw  [fill={rgb, 255:red, 139; green, 87; blue, 42 }  ,fill opacity=0.28 ] (92.66,222.38) .. controls (92.7,196.54) and (104.49,180.37) .. (119.01,186.26) .. controls (133.53,192.15) and (145.27,217.87) .. (145.24,243.71) .. controls (145.21,269.55) and (133.42,285.72) .. (118.9,279.84) .. controls (104.38,273.95) and (92.63,248.22) .. (92.66,222.38) -- cycle ;
\draw  [dash pattern={on 4.5pt off 4.5pt}]  (131,190) .. controls (170.6,160.3) and (110.23,121.78) .. (147.83,91.9) ;
\draw [shift={(149,91)}, rotate = 503.13] [color={rgb, 255:red, 0; green, 0; blue, 0 }  ][line width=0.75]    (10.93,-3.29) .. controls (6.95,-1.4) and (3.31,-0.3) .. (0,0) .. controls (3.31,0.3) and (6.95,1.4) .. (10.93,3.29)   ;
\draw  [dash pattern={on 4.5pt off 4.5pt}]  (142,211) .. controls (181.6,181.3) and (205.52,198.64) .. (254.51,175.71) ;
\draw [shift={(256,175)}, rotate = 514.36] [color={rgb, 255:red, 0; green, 0; blue, 0 }  ][line width=0.75]    (10.93,-3.29) .. controls (6.95,-1.4) and (3.31,-0.3) .. (0,0) .. controls (3.31,0.3) and (6.95,1.4) .. (10.93,3.29)   ;
\draw    (60,397) -- (419,396.01) ;
\draw [shift={(421,396)}, rotate = 539.8399999999999] [color={rgb, 255:red, 0; green, 0; blue, 0 }  ][line width=0.75]    (10.93,-3.29) .. controls (6.95,-1.4) and (3.31,-0.3) .. (0,0) .. controls (3.31,0.3) and (6.95,1.4) .. (10.93,3.29)   ;

\draw (87,285) node [anchor=north west][inner sep=0.75pt]   [align=left] {`The Godfather'};
\draw (61,125) node [anchor=north west][inner sep=0.75pt]   [align=left] {Best \\Picture};
\draw (299,215) node [anchor=north west][inner sep=0.75pt]   [align=left] {Marlon Brando (0.0)};
\draw (260,127) node [anchor=north west][inner sep=0.75pt]   [align=left] {\textcolor[rgb]{0,0,0}{'The Sting'}};
\draw (120,40) node [anchor=north west][inner sep=0.75pt]   [align=left] {`Forrest Gump'};
\draw (281,55) node [anchor=north west][inner sep=0.75pt]   [align=left] {Tom Hanks (0.0)};
\draw (287,343) node [anchor=north west][inner sep=0.75pt]   [align=left] {`Apocalypse Now' (0.05)};
\draw (102,92) node [anchor=north west][inner sep=0.75pt]   [align=left] {(0.2)};
\draw (203,150) node [anchor=north west][inner sep=0.75pt]   [align=left] {\textcolor[rgb]{0.25,0.46,0.02}{\textbf{(0.75)}}};
\draw (100.66,223.38) node [anchor=north west][inner sep=0.75pt]   [align=left] {\textbf{\textcolor[rgb]{0.25,0.46,0.02}{1972}}};

\end{tikzpicture}

%% file: tables/cronqs.tex
\begin{table}[h]
\resizebox{\columnwidth}{!}{%
\begin{tabular}{l|l}
\toprule
\multicolumn{1}{c|}{\textbf{Type}} & \multicolumn{1}{c}{\textbf{Example Questions}}   \\
\midrule
Simple Time & \textit{When did \{Stoke$\}_s$ have \{Tom Holford$\}_o$ in their team} \\
Simple Entity& \textit{Which movie won the \{Best Picture$\}_s$ in \{1973$\}_\tau$} \\
Before/After & \textit{Which movie won the \{Best Picture$\}_s$ after \{The Godfather$\}_o$} \\
First/Last & \textit{Name the award that \{Sydney Chapman$\}_s$ first received}  \\
Time Join & \textit{Name a teammate of \{Thierry Henry$\}_s$ in \{Arsenal$\}_o$}\\
\bottomrule
\end{tabular}%
}
\caption{Different types of temporal questions. $\{\cdot \}_s, \{\cdot \}_o, \{\cdot \}_\tau$  correspond to annotated entities $s,o \in \mathcal{E}$ and timestamps $\tau \in \mathcal{T}$.}
\label{tab:cronqs}
\end{table}

%% file: tables/main_results.tex
\begin{table*}[ht!]
\resizebox{0.8\textwidth}{!}{%
\begin{tabular}{l|r|rr|rr|r|rr|rr}
\toprule
\multicolumn{1}{c|}{\multirow{3}{*}{\textbf{Model}}} & \multicolumn{5}{c|}{\textbf{Hits@1}} & \multicolumn{5}{c}{\textbf{Hits@10}} \\ \cline{2-11} 
\multicolumn{1}{c|}{} & \multicolumn{1}{c|}{\multirow{2}{*}{\textbf{Overall}}} & \multicolumn{2}{c|}{\textbf{Question Type}} & \multicolumn{2}{c|}{\textbf{Answer Type}} & \multicolumn{1}{c|}{\multirow{2}{*}{\textbf{Overall}}} & \multicolumn{2}{c|}{\textbf{Question Type}} & \multicolumn{2}{c}{\textbf{Answer Type}} \\ \cline{3-6} \cline{8-11} 
\multicolumn{1}{c|}{} & \multicolumn{1}{c|}{} & \multicolumn{1}{c|}{\textbf{Complex}} & \multicolumn{1}{c|}{\textbf{Simple}} & \multicolumn{1}{c|}{\textbf{Entity}} & \multicolumn{1}{c|}{\textbf{Time}} & \multicolumn{1}{c|}{}& \multicolumn{1}{c|}{\textbf{Complex}} & \multicolumn{1}{c|}{\textbf{Simple}} & \multicolumn{1}{c|}{\textbf{Entity}} & \multicolumn{1}{c}{\textbf{Time}} \\
\midrule
BERT & 0.243 & 0.239 & 0.249 & 0.277 & 0.179  & 0.620 & 0.598 & 0.649 & 0.628 & 0.604 \\
RoBERTa & 0.225 & 0.217 & 0.237 & 0.251 & 0.177 & 0.585 & 0.542 & 0.644 & 0.583 & 0.591 \\
\hline
EmbedKGQA & 0.288 & 0.286 & 0.290 & { 0.411} & 0.057 & 0.672 & 0.632 & 0.725 & 0.850 & 0.341 \\
EaE & 0.288 & 0.257 & 0.329 & 0.318 & 0.231 & 0.678 & 0.623 & 0.753 & 0.688 & 0.698 \\
CronKGQA & 0.647 & 0.392 & 0.987 & 0.699 & 0.549 & 0.884 &  0.802 & 0.992 & 0.898 & 0.857 \\
\baseline & 0.745 & 0.562 & 0.990 & 0.831 & 0.585 & 0.944 & 0.906 & 0.993 & 0.962 & 0.910 \\
\hline
\methods{} & 0.799 & 0.655 & 0.990 & 0.876 & 0.653 &  0.957 & 0.930 & 0.993 & 0.972 & 0.929\\
\methodh{} & 0.918 & 0.864 & 0.990 & 0.926 & 0.903 &  0.978 & 0.967 & 0.993 & 0.980 & 0.974\\
\bottomrule
\end{tabular}%
}
\caption{Comparison against other methods.}
\label{tab:main-results}
\end{table*}


%% file: figures/random_times.tex
\definecolor{col1}{rgb}{0.60, 0.31, 0.64}
\definecolor{col2}{rgb}{0.30, 0.69, 0.29}
\definecolor{col3}{rgb}{0.22, 0.49, 0.72}
\definecolor{col4}{rgb}{0.89, 0.10, 0.11}
\definecolor{col5}{rgb}{1, 1, 0.8}

\begin{tikzpicture}
\tikzstyle{every node}=[font=\Huge]
\begin{axis}[legend style={at={(.9,0.9),anchor=north east}},
             legend style={legend pos=outer north east,},  title={Before/After},ylabel={Hits@1 (\%)}, every axis plot/.append style={ultra thick}, ymax=100, ymin=47
]

\addplot[mark=square*,col1] coordinates {
    (0.,71.4)
    (0.2,63.4)
    (0.5,57.0)
    (0.8, 54.7)

};

\addplot[mark=diamond*,col2] coordinates {
    (0, 67.0)
    (0.8, 67.0)
};
\addplot[mark=triangle*,col3] coordinates {
    (0, 54.0)
    (0.8, 54.0)
};
\end{axis}

\begin{axis}[legend style={at={(.9,0.9),anchor=north east}},
             legend style={legend pos=outer north east,}, title={First/Last},every axis plot/.append style={ultra thick}, xshift=9cm,ymax=100, ymin=47
]

\addplot[mark=square*,col1] coordinates {
    (0.0,85.3)
    (0.2,64.5)
    (0.5,51.4)
    (0.8, 49.7)

};

\addplot[mark=diamond*,col2] coordinates {
    (0.0,57.0)
    (0.8,57.0)
    
};

\addplot[mark=triangle*,col3] coordinates {
    (0, 49.3)
    (0.8, 49.3)
    
};

\end{axis}

\begin{axis}[legend columns=-1,legend style={at={(0.5,1.35),anchor=north east,}},
              title={Time Join}, every axis plot/.append style={ultra thick}, xshift=18cm, legend style={/tikz/every even column/.append style={column sep=0.5cm}},ymax=100, ymin=47
]

\addplot[mark=square*,col1] coordinates {
    (0.0,97.8)
    (0.2,92.9)
    (0.5,84.3)
    (0.8, 83.6)

};
\addlegendentry{\methodh}

\addplot[mark=diamond*,col2] coordinates {
    (0.0,89.4)
    (0.8,89.4)
    
};
\addlegendentry{\methods}

\addplot[mark=triangle*,col3] coordinates {
    (0,83.3)
    (0.8,83.3)
    
};
\addlegendentry{\baseline}

\addplot[mark=otimes*,col4] coordinates {
    (0.8,0)
    
};
\addlegendentry{CronKGQA}
\end{axis}






\end{tikzpicture}

%% file: figures/missing_link.tex
\definecolor{col1}{rgb}{0.60, 0.31, 0.64}
\definecolor{col2}{rgb}{0.30, 0.69, 0.29}
\definecolor{col3}{rgb}{0.22, 0.49, 0.72}
\definecolor{col4}{rgb}{0.89, 0.10, 0.11}
\definecolor{col5}{rgb}{1, 1, 0.8}

\begin{tikzpicture}
\tikzstyle{every node}=[font=\Huge]
\begin{axis}[legend style={at={(.9,0.9),anchor=north east}},
             legend style={legend pos=outer north east,},  title={Before/After},ylabel={Hits@1 (\%)}, every axis plot/.append style={ultra thick}, ymin=20, ymax=100
]

\addplot[mark=square*,col1] coordinates {
    (0.,71.4)
    (0.2,53.9)
    (0.33, 53.3)
    (0.5,44.4)

};

\addplot[mark=diamond*,col2] coordinates {
    (0.,67.0)
    (0.2,49.9)
    (0.33, 46.8)
    (0.5,43.1)

};
\addplot[mark=triangle*,col3] coordinates {
    (0.,54.0)
    (0.2, 46.7)
    (0.33, 44.5)
    (0.5, 42.1)

};

\addplot[mark=otimes*,col4] coordinates {
    (0.,28.8)
    (0.2,24.5)
    (0.33,25.7)
    (0.5,25.2)

};

\end{axis}
\begin{axis}[legend style={at={(.9,0.9),anchor=north east}},
             legend style={legend pos=outer north east,}, title={First/Last}, every axis plot/.append style={ultra thick},ymin=20, ymax=100, xshift=9cm
]

\addplot[mark=square*,col1] coordinates {
    (0.0,85.3)
    (0.2,64.0)
    (0.33, 80.3)
    (0.5,47.4)
    
};

\addplot[mark=diamond*,col2] coordinates {
    (0.0,57.0)
    (0.2,48.4)
    (0.33, 44.4)
    (0.5,41.4)
    
};
\addplot[mark=triangle*,col3] coordinates {
    (0.0, 49.3)
    (0.2, 45.0)
    (0.33, 44.1)
    (0.5, 40.0)

};

\addplot[mark=otimes*,col4] coordinates {
    (0.0,37.1)
    (0.2,36.2)
    (0.33, 37.0)
    (0.5,36.0)
    
};
\end{axis}
\begin{axis}[legend style={at={(.9,0.9),anchor=north east}},
             legend style={legend pos=outer north east,}, title={Time Join}, every axis plot/.append style={ultra thick},ymin=20, ymax=100, xshift=18cm
]

\addplot[mark=square*,col1] coordinates {
    (0.0,97.8)
    (0.2,78.5)
    (0.33, 82.6)
    (0.5,60.9)
    
};
\addplot[mark=diamond*,col2] coordinates {
    (0.0,89.4)
    (0.2,73.8)
    (0.33, 65.6)
    (0.5,57.6)
    
};
\addplot[mark=triangle*,col3] coordinates {
    (0.0,83.3)
    (0.2, 67.6)
    (0.33, 44.5)
    (0.5, 53.5)

};
\addplot[mark=otimes*,col4] coordinates {
    (0.0,51.1)
    (0.2,38.4)
    (0.33, 35.9)
    (0.5,33.3)
    
};

\end{axis}
\begin{axis}[legend columns=-1,legend style={at={(-0.5,1.35),anchor=north east,}},
             title={Simple Entity/Time},  every axis plot/.append style={ultra thick}, ymin=60, ymax=100, xshift=27cm, legend style={/tikz/every even column/.append style={column sep=0.5cm}}
]
\addplot[mark=diamond*,col1] coordinates {
    (0.0,99.0)
    (0.2,81.0)
    (0.33, 84.2)
    (0.5,64.8)
    
};

\addplot[mark=diamond*,col2] coordinates {
    (0.0,99.0)
    (0.2,79.5)
    (0.33, 75.4)
    (0.5,63.3)
    
};

\addplot[mark=diamond*,col3] coordinates {
    (0.0,99.0)
    (0.2, 80.2)
    (0.33, 77.3)
    (0.5, 65.1)
    
};

\addplot[mark=diamond*,col4] coordinates {
    (0.0,98.7)
    (0.2,76.1)
    (0.33, 72.1)
    (0.5,61.1)
    
};

\end{axis}
\end{tikzpicture}

%% file: tables/hard_soft.tex
\begin{table}[h]
\centering
\resizebox{0.8\textwidth}{!}{%
\begin{tabular}{l|ccc|r}
\toprule
& \multicolumn{3}{c}{\textbf{Complex Questions}} & \textbf{All} \\
              & Before/ & First/ & Time & \\
              & After & Last & Join \\ 
\midrule

CronKGQA &  0.256 & 0.371 & 0.511 & 0.647 \\
\baseline{} &  0.540 & 0.493 & 0.833 & 0.745 \\
\hline
CronKGQA-Soft &  0.341 & 0.375 & 0.671 & 0.672 \\
\baseline-Soft & 0.430 & 0.468 & 0.766  &  0.708 \\
\methods & 0.670 &  0.570 & 0.894 &  0.799 \\
\hline
CronKGQA-Hard & 0.379 & 0.445 & 0.942  &  0.728 \\
\baseline-Hard  & 0.442 & 0.470 & 0.955 & 0.748 \\
\methodh  & 0.714 & 0.853 & 0.978 &  0.918\\
\bottomrule
\end{tabular}%
}
\caption{Hits@1 for different complex type questions. }
\label{tab:hard-soft}

\end{table}



%% file: figures/before_after.tex
\definecolor{col1}{rgb}{0.60, 0.31, 0.64}
\definecolor{col2}{rgb}{0.30, 0.69, 0.29}
\definecolor{col3}{rgb}{0.22, 0.49, 0.72}
\definecolor{col4}{rgb}{0.89, 0.10, 0.11}
\definecolor{col5}{rgb}{1, 1, 0.8}

\begin{tikzpicture}
\tikzstyle{every node}=[font=\Huge]

\begin{axis}[legend style={at={(.9,0.9),anchor=north east}},
             legend style={legend pos=outer north east,},  title={Before \& After},ylabel={Hits@$k$ (\%)}, every axis plot/.append style={ultra thick}, ymin=0, ymax=61
]
\addplot[mark=square*,col1] coordinates {
    (1, 10.7)
    (2, 19.2)
    (5, 45.2)
    (10, 58.6)
};

\addplot[mark=diamond*,col2] coordinates {
    (1,11.5)
    (2, 22.6)
    (5, 46.0)
    (10, 60.5)
};

\addplot[mark=triangle*,col3] coordinates {
    (1, 6.1)
    (2, 10.3)
    (5, 27.6)
    (10, 45.6)
};

\addplot[mark=diamond*,col4] coordinates {
    (1, 1.1)
    (2, 2.3)
    (5, 4.2)
    (10, 6.5)
};
\end{axis}

\begin{axis}[legend columns=-1,legend style={at={(1,1.35),anchor=north east,}},  title={Before/After \& First/Last},  every axis plot/.append style={ultra thick}, ymin=0, ymax=61, xshift=10cm
]

\addplot[mark=square*,col1] coordinates {
    (1,11.1)
    (2, 18.0)
    (5, 26.4)
    (10, 28.7)
};
\addlegendentry{\methodh} 

\addplot[mark=diamond*,col2] coordinates {
    (1, 12.3)
    (2, 22.2)
    (5, 33.0)
    (10, 41.4)
};
\addlegendentry{\methods} 

\addplot[mark=triangle*,col3] coordinates {
    (1, 5.4)
    (2, 12.6)
    (5, 19.5)
    (10, 23.0)
};
\addlegendentry{\baseline} 
\addplot[mark=otimes*,col4] coordinates {
    (1, 1.9)
    (2, 5.7)
    (5, 8.0)
    (10, 11.1)
};
\addlegendentry{CronKGQA} 
\end{axis}
\end{tikzpicture}

%% file: appendix.tex
\section{Appendix}

\subsection{CronQuestions Statistics}
Additional statistics of the CronQuestions dataset are shown in Table~\ref{tab:dataset-stats-questions}. Simple Reasoning questions contains Simple Entity and Simple Time questions, while Complex Reasoning contain Before/After, First/Last, and Time Join questions. Entity Answer questions contain Simple Entity, Before/After, First/Last, and Time Join questions, while Time Answer questions contain Simple Time and First/Last.

\begin{table}[h]
\resizebox{0.7\columnwidth}{!}{%
\begin{tabular}{l|rrr}
\hline
               & \multicolumn{1}{l}{\textbf{Train}} & \multicolumn{1}{l}{\textbf{Dev}} & \multicolumn{1}{l}{\textbf{Test}} \\ \hline
Simple Entity  & 90,651                             & 7,745                            & 7,812                             \\
Simple Time    & 61,471                             & 5,197                            & 5,046                             \\
Before/After   & 23,869                             & 1,982                            & 2,151                             \\
First/Last     & 118,556                            & 11,198                           & 11,159                            \\
Time Join      & 55,453                             & 3,878                            & 3,832                             \\ \hline
Simple Reasoning  & 152,122                            & 12,942                           & 12,858
\\
Complex Reasoning & 197,878                            & 17,058                           & 17,142                            \\ \hline

Entity Answer  & 225,672                            & 19,362                           & 19,524                            \\
Time Answer    & 124,328                            & 10,638                           & 10,476                            \\ \hline
\textbf{Total} & \multicolumn{1}{l}{350,000}        & \multicolumn{1}{l}{30,000}       & \multicolumn{1}{l}{30,000}        \\ \hline
\end{tabular}
}
\caption{Number of questions in our dataset across different types of reasoning required and different answer types.}
\label{tab:dataset-stats-questions}
\end{table}

\subsection{Generation of Additional Complex Questions}

We create `before \& after' questions that include both `before' and `after' constraints, e.g., ``\textit{Which movies won The Best Picture \underline{after} The Godfather and \underline{before} Forrest Gump?}". Additionally, we create `before/after \& first/last' questions that include both `before/after' and `first/last' constraints, e.g., ``\textit{Which is the \underline{last} movie to win Best Picture \underline{after} The Godfather?}". For both new complex types, we end up with 261 QA pairs that are only used for inference (unseen during training). We also provide the QA pairs in the submission files.

The two evaluation datasets were created as follows. The `before \& after' dataset was generated by identifying all before/after questions in the original CronQuestions dataset which contained the `position held' relation. These questions were then filtered to a subset derived from 14 templates which could be easily modified to contain both before and after conditions while remaining syntactically correct. For example, the template `Who held the position of {tail} after {head}' was included in this set because it can be converted into a before \& after question by adding ` and before {tail2}' to the end of the sentence. All of the new before \& after questions were created in this manner by adding either ` and before {tail2}' or ` and after {tail2}' for questions which originally contained after or before conditions (respectively).

The `{tail2}' and answer entities for these questions were selected by first identifying all answer entities from the original question which have a start time in the TKG which is strictly greater (less) than that of the start time of the 1{head}' entity for questions containing an after (before) condition. These entities were then sorted in ascending order of their closeness in time to the 1{head}' entity. The entity containing the third-closest distinct start time to that of the `{head}' entity was selected as the `{tail2}' entity while all other entities containing a start date between that of the `{head}' and `{tail2}' entities were used as answer entities. 

The `before/after \& first/last' questions were generated using a similar process. For questions originally containing an `after' condition, we also add the word  `first' prior to the `{tail}' relation. Likewise, we add the word `last' prior to the `{tail}' relation for questions which originally contained a `before' condition. The answer set is then filtered to only contain the entity which has the closest start date in time to that of the `{head}' entity. 

\subsection{Variants of \method}
\textbf{Difference with TComplEx}. TComplEx is a TKG embedding scoring function (\textit{decoder}) and can only answer questions that involve a \textit{single} fact, e.g., $(s,r,?,\tau)$, with a \textit{provided} TKG relation $r$. TempoQR handles questions that involve \textit{multiple} facts and does not rely on a specific relation $r$. The main contribution here is to \textit{encode} semantic and temporal information from multiple facts to a single question representation $q$, which can be used as a \textit{virtual} relation (Sec. 4.2 and 4.3). For scoring answers using $q$, different decoders can be implemented via Eq.(9) and (10), e.g., the TComplEx or a dot product decoder. For TempoQR-Soft, Eq.(6) should be adapted according to the decoder. 

We also include the following Table to directly compare TComplEx with TempoQR and to evaluate the performance with other decoders for complex temporal questions (H@1). The TComplEx model is implemented via Eq.(1) by providing the ground-truth relation $r$ of each question.  The first row of the Table shows that TempoQR vastly outperforms TComplEx since the latter cannot handle complex questions. Moreover, the second row shows that our method is competitive across different decoders, such as the dot-product $\phi_{\text{dot}}$.
\resizebox{0.95\columnwidth}{!}{
\centering
\begin{tabular}{|c|c|c|c|c|} 
 \hline
 Model$\rightarrow$&TComplEx&CronKGQA&TempoQR-Soft&TempoQR-Hard
\\ 
Decoder$\downarrow$& & & & \\
\hline
  $\phi_{\text{TComplEx}}$ & 0.022 & 0.392 & 0.655 & 0.864 \\
  \hline
 $\phi_{\text{dot}}$ & Not applicable & 0.248 & 0.366 & 0.721 \\ 
 \hline
\end{tabular}
}

\textbf{Ensemble of Hard and Soft Supervision}. When we add time embeddings of both approaches and measure the H@1(\%) for complex questions, the ensemble improves by 2.1 points over TempoQR-Hard for `before/after' questions where TempoQR-Hard cannot retrieve the accurate time from the TKG. For `first/last' and `time join' question, the soft supervision introduces noise and actually decreases the performance compared to TempoQR-Hard by 1.9 and 6.7 points, respectively.

\subsection{Fusion of Time Embeddings}

\textbf{\method-cat and \method-att}. \method-cat and \method-att are variations of \method{} with different fusion implementations compared to the one in Section~\ref{sec:fusion}. In \method-cat, we substitute the sum operator in ~\eqref{eq:tempfusion} $\big(\mathbf{q}_{T_i} = \mathbf{q}_{T_i} +\mathbf{t}_1 +\mathbf{t}_2 \big) $ by concatenating the entity and time embeddings followed by a learnable projection. In \method-att, we append the time embeddings to the embedding matrix  as new tokens of the sentence. This is equivalent to transforming the given sentence from, e.g., \textit{“Which movie won the Best Picture after The Godfather”} to \textit{“Which movie won the Best Picture after The Godfather, 1972”} and applying \baseline{}. We use \method-att with hard-supervision only.

\input{tables/fusion}
Table~\ref{tab:fusion} shows the results for different strategies of fusing time embeddings with hard or soft supervision. As we can see, our implementation of summing entity and time embeddings (\method) performs the best. We note that this is typically the way transformer-based architectures combine token and positional embeddings. Moreover, \method-cat and \method-att perform much worse for `time join' and `simple' questions. This means that a lot of important time information is ignored by their projection and attention layer, respectively. Whereas, by summing the time embeddings, we ensure that all the information is encoded.

\subsection{Training Size}

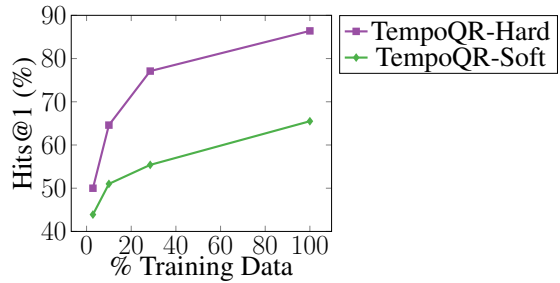
\begin{figure}[h]
\centering
\resizebox{0.9\columnwidth}{!}{\input{figures/train_split}}
\caption{Evaluation for different training size. }
\label{fig:train}
\end{figure}

Figure~\ref{fig:train} shows the performance for complex questions for different training  sizes. As the training size reduces, both \methodh and \methods perform worse. We observe a slightly faster performance drop for \methodh.

\subsection{Is Time All We Need?}
In this experimental configuration, the goal is to evaluate \methodh with different ways that generate time embeddings. First, we use TComplEx embeddings for the (i) start and end timestamps (Section~\ref{sec:TAQE}) (TComplEx-Start\&End) and (ii) a randomly sampled timestamp between the start and end timestamp (TComplEx-Sampled). The second approach is instead of using TComplEx embeddings to use positional embeddings as in~\cite{vaswani2017transformer}. The positional embedding are generated based on each timestamp's identifier and are ensured to  preserve the ordering of the timestamps in the embedding space. We use positional embeddings of the start and end timestamp for \methodh (PosEmb-Start\&End). Finally, we generate completely random time embeddings, which we use them for the start and end timestamps (RandEmb-Start\&End). This ensures that all time embeddings are well-separated. In all cases, the time embeddings are not updated during QA.

\input{tables/hard_te}

Table~\ref{tab:hard-te} shows the results for differently generated time embeddings for \methodh. As we can see, using the start and end timestamps is a key decision that provides the best results (TComplEx-Sampled has the lowest performance). Table~\ref{tab:hard-te} indicates is that the time embeddings themselves do not play an important role. What \methodh needs is to be able to associate given questions with different times. This is confirmed since RandEmb have the perform equally with TComplEx embeddings. Here, the random embeddings are nothing more than a unique identifier for each timestamp. They also perform the best for `first/last' questions, since the timestamps are well-separated in the embedding space. On the other hand, PosEmb perform the best for `before/after' questions. This happens because the timestamp ordering is preserved in the embedding space and allows for better `before/after' reasoning.

%% file: tables/fusion.tex
\begin{table}[h]
\centering
\resizebox{0.9\columnwidth}{!}{%
\begin{tabular}{l|ccccc}
\toprule
& \multicolumn{3}{c}{\textbf{Complex Questions}} & \multicolumn{2}{c}{\textbf{Simple Questions}}  \\
              & Before/ & First/ & Time & Simple & Simple\\
              & After & Last & Join & Entity & Time \\ 
\midrule
\textbf{Soft Super.} & & & & & \\
\method{}-cat & 0.605 & 0.518& 0.794 & 0.985 & 0.867\\
\method{}  & 0.670 &  0.570 & 0.894  & 0.991 & 0.987\\
\hline
\textbf{Hard Super.} & & & & &\\
\method{}-cat & 0.692 & 0.847& 0.866 & 0.989 & 0.940 \\
\method{}-att & 0.683 & 0.835 & 0.871 & 0.989 & 0.881 \\
\method{}  & 0.714 & 0.853 & 0.978 & 0.992 & 0.988 \\
\bottomrule
\end{tabular}%
}
\caption{Hits@1 for fusion strategies. }
\label{tab:fusion}
\end{table}

%% file: figures/train_split.tex
\definecolor{col1}{rgb}{0.60, 0.31, 0.64}
\definecolor{col2}{rgb}{0.30, 0.69, 0.29}
\definecolor{col3}{rgb}{0.22, 0.49, 0.72}
\definecolor{col4}{rgb}{0.89, 0.10, 0.11}
\definecolor{col5}{rgb}{1, 1, 0.8}

\begin{tikzpicture}
\tikzstyle{every node}=[font=\Huge]
\begin{axis}[ legend style={legend pos=outer north east,}, every axis plot/.append style={ultra thick}, ymin=40, ymax=90, ylabel={Hits@1 (\%)}, xlabel= \% Training Data
]
\addplot[mark=square*,col1] coordinates {
    (2.85, 50.0)
    (10, 64.6)
    (28.5, 77.1)
    (100, 86.4)
};
\addlegendentry{\methodh}

\addplot[mark=diamond*,col2] coordinates {
    (2.85,43.9)
    (10, 51.0)
    (28.5, 55.4)
    (100, 65.5)
};
\addlegendentry{\methods}

\end{axis}
\end{tikzpicture}

%% file: tables/hard_te.tex
\begin{table}[h]
\centering
\resizebox{0.75\columnwidth}{!}{%
\begin{tabular}{l|ccc}
\toprule
\textbf{\methodh}  & Before/ & First/ & Time \\
\textbf{Time Embeddings}              & After & Last & Join  \\ 
\midrule
TComplEx-Sampled & 0.602 & 0.516 & 0.919 \\
TComplEx-Start\&End & 0.714 & 0.853 & 0.978  \\
PosEmb-Start\&End & 0.724 &  0.832 &  0.976 \\
RandEmb-Start\&End & 0.713 & 0.868  &  0.976  \\
\bottomrule
\end{tabular}%
}
\caption{Hits@1 for different time embeddings of \methodh. }
\label{tab:hard-te}
\end{table}